\documentclass[11pt]{article}
\usepackage{paclic35}
\usepackage{times}
\usepackage{latexsym}
\usepackage{amsmath}
\usepackage{multirow}
\usepackage{url}

\usepackage[utf8]{inputenc}
\usepackage{CJK}
\usepackage[T1]{fontenc}
\usepackage{array}
\usepackage{booktabs}
\usepackage{enumitem}
\usepackage{microtype}
\usepackage{supertabular}
\usepackage{longtable}
\usepackage{color}
\usepackage{subfigure}
\usepackage{amsfonts,amssymb} 
\usepackage{graphicx}
\usepackage{bm}
\usepackage{amsmath}
\usepackage{multicol}
\usepackage{multirow}
\usepackage{makecell}
\usepackage{verbatim}

\title{Multi-tasking Dialogue Comprehension with Discourse Parsing}
\author{Yuchen He\textsuperscript{1,2,3,\#}, Zhuosheng Zhang\textsuperscript{1,2,3,\#}, Hai Zhao\textsuperscript{1,2,3,\thanks{\ \ Corresponding author. \# Equal contribution. This paper was partially supported by Key Projects of National Natural Science Foundation of China (U1836222 and 61733011).}}\\
\textsuperscript{1} Department of Computer Science and Engineering, Shanghai Jiao Tong University\\
\textsuperscript{2} Key Laboratory of Shanghai Education Commission for Intelligent Interaction\\
and Cognitive Engineering, Shanghai Jiao Tong University\\
\textsuperscript{3}MoE Key Lab of Artificial Intelligence, AI Institute, Shanghai Jiao Tong University\\
\texttt{\{21644h,zhangzs\}@sjtu.edu.cn,zhaohai@cs.sjtu.edu.cn}\\
}

\date{}

\begin{document}
\maketitle
\begin{abstract}
Multi-party dialogue machine reading comprehension (MRC) raises an even more challenging understanding goal on dialogue with more than two involved speakers, compared with the traditional plain passage style MRC. To accurately perform the question-answering (QA) task according to such multi-party dialogue, models have to handle fundamentally different discourse relationships from common non-dialogue plain text, where discourse relations are supposed to connect two far apart utterances in a linguistics-motivated way.
To further explore the role of such unusual discourse structure on the correlated QA task in terms of MRC, we propose the first multi-task model for jointly performing QA and discourse parsing (DP) on the multi-party dialogue MRC task. Our proposed model is evaluated on the latest benchmark Molweni, whose results indicate that training with complementary tasks indeed benefits not only QA task, but also DP task itself. We further find that the joint model is distinctly stronger when handling longer dialogues which again verifies the necessity of DP in the related MRC.
\end{abstract}

\section{Introduction}\label{sec:introduction}
Machine reading comprehension (MRC) is essentially formed as a question-answering (QA) task subject to a given context like passages \cite{hermann2015teaching,rajpurkar2016squad}. Recently, more and more attention is raised on a special MRC type whose given context is a dialogue text \cite{reddy2019coqa,choi2018quac}. Training machines to understand dialogue has been shown more challenging than the common MRC as every utterance in dialogue has an additional property of speaker role, which breaks the continuity as that in common non-dialogue texts due to the presence of crossing dependencies which are commonplace in multi-party chat \cite{allen1994trains,perez2017dialog}. Thus dialogue demonstrates quite a different discourse relationship mode from the non-dialogue in that consecutive utterances usually have a type of discourse relation \cite{afantenos2015discourse,shi2019deep,li-etal-2020-molweni,DBLP:journals/corr/abs-2104-12377}. Recently, there emerges an even more challenging dialogue MRC task, the multi-party one, which involves more than two speakers in the given dialogue passage \cite{li-etal-2020-molweni,DBLP:journals/corr/abs-2104-12377} and further demonstrates unusual discourse structures such as quite a lot of adjacent utterances do not have any semantic relationship. Being harder for comprehension, multi-party dialogue MRC has a great value of application which can be applied to frontiers such as intelligent human-computer interface and knowledge graph building.

\begin{figure}
		\centering
		\includegraphics[width=0.46\textwidth]{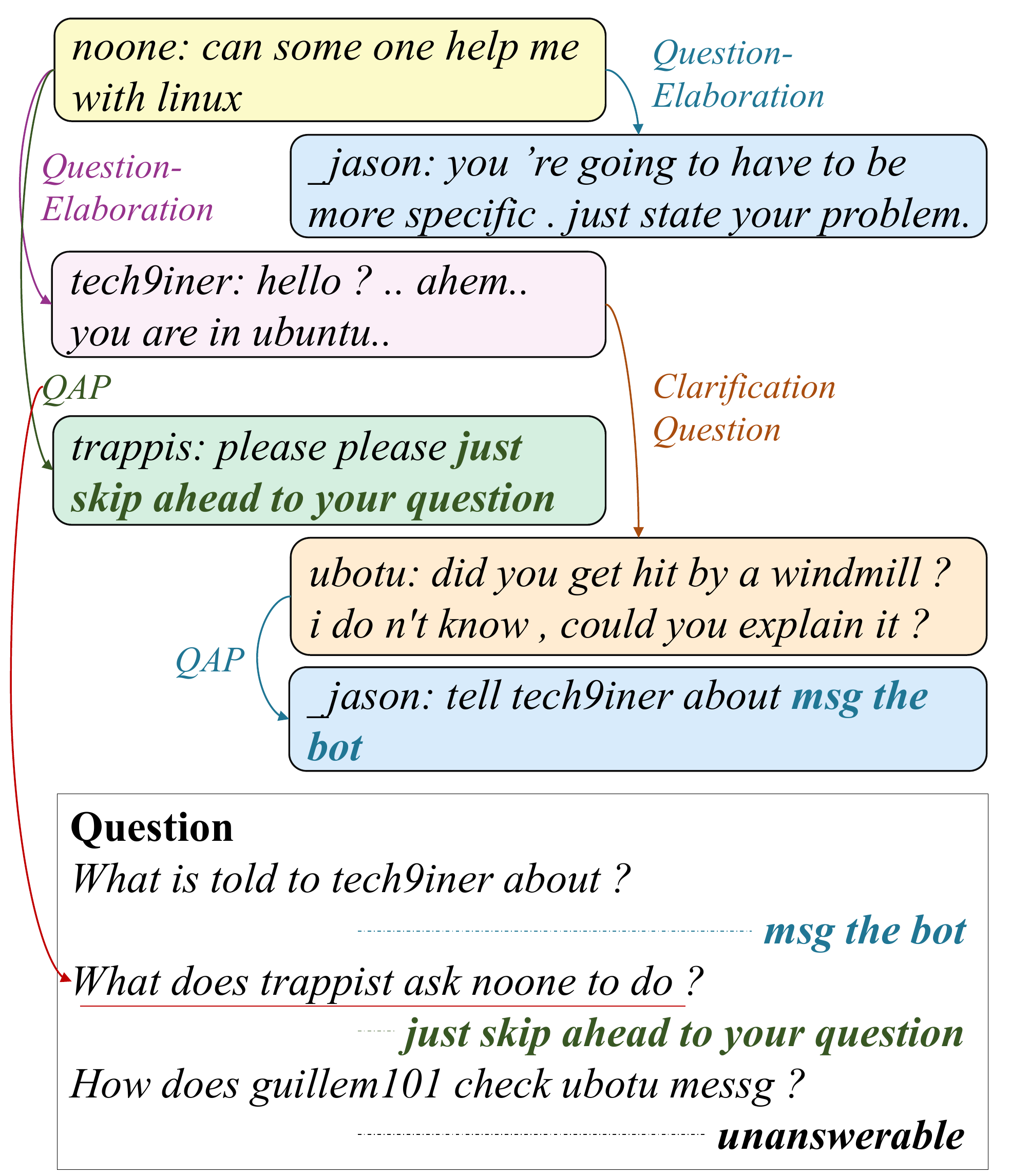}
		\caption{\label{fig:introduction example}An example of multi-party dialogue MRC in the Molweni dataset \cite{li-etal-2020-molweni}.}
\end{figure}
As shown in Figure \ref{fig:introduction example}, our work tries to extract the answer to the given question from the multi-party dialogue. Unlike texts used in typical MRC tasks, the multi-party dialogue has manifold sentence patterns and the topics of adjacent sentences can be totally irrelevant sometimes. The context of multi-party dialogues is defined by abstract discourse structures rather than sentence positions. 


Considering that the question-answering (QA) and discourse parsing (DP) tasks in the multi-party dialogue MRC are correlated and share close relations, it is supposed to naturally model these two tasks as one. Intuitively, the discourse structure entailed in the DP task would be helpful for modeling the inner utterance relationships in the dialogue context. For example, as Figure \ref{fig:introduction example} shows, the first and fourth utterances is a question-answering pair (QAP) (which is marked by the red arrow) which helps strengthen the connection between the two utterances and might help answer the second question. Meanwhile, the QA task aims to extract the salient span-level answers that potentially benefit DP. However, it is surprising such a model design does not appear until this work makes the first attempt by doing so. 

In this work, we present a unified model for the multi-party dialogue MRC, which for the first time formally integrates such two diverse tasks for one purpose in multi-task learning (MTL) mode. We expect the model can deal with both QA and DP subtasks well and perform better than in individual tasks. By carefully selecting a proper testbed, our proposed method will be evaluated on the latest multi-party dialogue MRC benchmark, Molweni \cite{li-etal-2020-molweni}, which both tasks can exploit accurate human annotations, to guarantee the reliability of our results. Experimental results indicate that multi-tasking the complementary tasks indeed benefits not only QA task, but also DP task itself. We further find that the joint model performs better when handling longer dialogues, which proves the strong correlations between the two tasks. As a result, our model also achieves state-of-the-art results on the Molweni multi-party dialogue dataset.

\section{Background and Related Work}\label{sec:related work}

\subsection{QA-based MRC}\label{subsec:MRC DP}
MRC task aims at teaching the machine to answer questions according to given reference texts \cite{hermann2015teaching,rajpurkar2016squad,zhang2020machine}. The study of MRC has experienced two significant peaks, namely, 1) the burst of deep neural networks \cite{wei2018fast,seo2018bidirectional}; 2) the evolution of pre-trained language models (PrLMs) \cite{devlin-etal-2019-bert,Clark2020ELECTRA:}. In the early stage, MRC was regarded as the form of triple-style (passage, question, answer) question answering (QA) task, such as the cloze-style \cite{hermann2015teaching,hill2015goldilocks}, multiple-choice \cite{lai2017race,sun2019dream}, and span-QA \cite{rajpurkar2016squad,Rajpurkar2018Know}. Among these types, span-based QA MRC has aroused the most research interests. 

Recently, more and more attention is raised on a special MRC type whose given passage is a dialogue text \cite{reddy2019coqa,choi2018quac}. In this work, we deal with the QA-based MRC task on multi-party dialogues, which requires the machine to extract a consecutive piece from the original dialogue. Multi-party dialogue comprehension involves more than two speakers, and there is a complicated phenomenon of crossing dependencies in multi-party dialogues. It has been shown much more challenging than the traditional MRC models \cite{li-etal-2020-molweni} due to the requirement to handle quite different discourse relationship modes from common non-dialogue plain text, where discourse relations may quite possibly connect two far apart utterances.



\subsection{Discourse Parsing}
Discourse parsing focuses on the discourse structure and relationships of texts, whose aim is to predict the relations between discourse units so as to disclose the discourse structure between those units. Discourse parsing has been studied by researchers especially in linguistics for decades. Previous studies have shown that discourse structures are beneficial for various natural language processing (NLP) tasks, including dialogue understanding \cite{asher2016discourse,takanobu2018weakly,gao-etal-2020-discern,jia2020multi}, question answering \cite{chai2004discourse,verberne2007evaluating,mihaylov2019discourse}, and sentiment analysis \cite{cambria2013new,nejat2017exploring}.

Most of the previous works for discourse parsing (DP) are based on the linguistic discourse datasets, such as Penn Discourse TreeBank (PDTB) \cite{prasad2008penn} and Rhetorical Structure Theory Discourse TreeBank (RST-DT) \cite{mann1988rhetorical}. PDTB focuses on shallow discourse relations but ignores the overall discourse structure \cite{qin2017adversarial,cai2017pair,bai2018deep,yang2018scidtb}.
In contrast, RST is constituency-based, where related adjacent discourse units are merged to form larger units recursively \cite{braud2017cross,wang2017two,yu2018transition,joty2015codra,li2016discourse,liu2017learning}.
Compared with the traditional DP tasks which are linguistically motivated, our work is application-driven from dialogue comprehension scenarios and devotes itself to handling the multiparty dialogues that involve more complex utterance relationships and speaker role transitions. However, most of the previous constituency-based DP tasks only focus on plain texts and does not allow non-adjacent relations, which makes it inapplicable for modeling multi-party dialogues. In terms of serving such a purpose, we are the first to present a  pre-trained language model (e.g., BERT \cite{devlin-etal-2019-bert}) based method for discourse parsing to our best knowledge.

\begin{figure*}[ht]
		\centering
		\includegraphics[width=\textwidth]{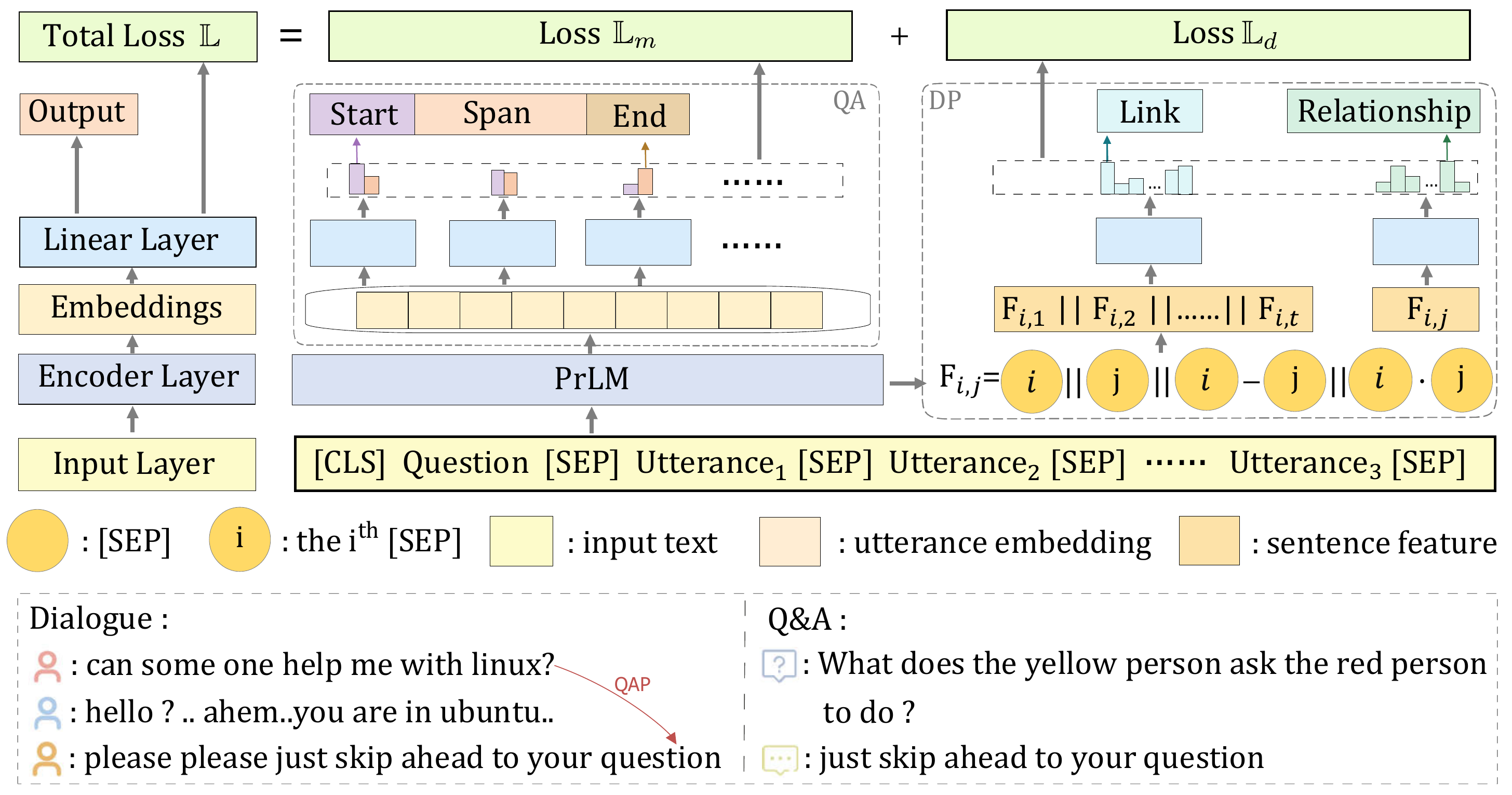}
		\caption{\label{fig:PrLM}The overview of the joint model. The top half part is the PrLM. The left lower part is the QA model, and the right lower part is the DP model.}
\end{figure*}

\section{Methods}\label{sec:methods}
\subsection{Feature Extraction}\label{subsec:embeddings}
Figure \ref{fig:PrLM} overviews our multi-party dialogue MRC model which parallelly includes modules of QA and DP. We apply PrLMs to encode our dialogue context and questions. Before data input, we first append padding symbols to fill the content for texts with tokens less than the preset value and add separators ($\texttt{[CLS]}$ and $\texttt{[SEP]}$) between question and dialogue or adjacent utterances, following the standard process of using PrLMs \cite{devlin-etal-2019-bert}. The positions of separators in the dialogue will be recorded to separate single utterance information for further DP task. We put the question in front of the dialogue to take full advantage of the knowledge learned in the next sentence prediction task of the pre-training stage and get abundant semantic information of the question. We concatenate the question $\boldsymbol{Q}=w_q^1w_q^2...w_q^n$ and dialogue context $\boldsymbol{D}=w_d^1w_d^2...w_d^m$ as a whole to feed the PrLM encoder and get the output text feature:
$
\boldsymbol{S}$=$\textrm{encode}([CLS],\boldsymbol{Q},[SEP],\boldsymbol{D},[SEP]),
$
where $\boldsymbol{S}$ is the contextualized sequence representations, and $w_q^i$ ($1\leq i\leq n$) and $w_d^j$ ($1\leq j\leq m$) represent tokens of texts. Variables $n$ and $m$ respectively mean the number of tokens in the question and dialogue. 

The output feature can be used in QA task directly, but for DP task, we have to do further processing to get the eigenvectors that represent the utterance relationship. After obtaining the features, we fetch the vectors at corresponding positions of separators to represent the utterances respectively. On the grounds of Euclidean and cosine distance and considering the asymmetry of utterance relationship, we use this cascade as the relationship feature to do DP task as Figure \ref{fig:PrLM} shows:
$
F_{i,j}= (E_{SEP}^i, E_{SEP}^j, E_{SEP}^i-E_{SEP}^j, E_{SEP}^i\cdot E_{SEP}^j),
$
where $E_{SEP}^i$ is the output feature of the $i^{th}$ separator in the dialogue for the $i^{th}$ utterance.
\subsection{Prediction}\label{subsec:answer pred}
For the QA task, we treat question answering as a multi-classification task by using fully connected layers to predict the start logits and end logits of the answer over the given dialogue. Then the most likely start and end positions are computed by using softmax as an actived function and the answer piece is extracted from the initial dialogue. Take the prediction of the start position for example:
$P_s=\textrm{argmax}(\textrm{softmax}(\boldsymbol{W}_{s}\boldsymbol{S})),$ where $P_s$ is the predicted start position, $\boldsymbol{W}_{s}$ is the weight matrix and $\boldsymbol{S}$ is the text feature. It is important to note that in this work, we need to deal with unanswerable questions. A score of the most likely answer span will be calculated and compared to a no-answer score to determine whether the question is answerable \cite{zhang2020retrospective}.

For the DP task, we represent the relationships of utterances by dependency trees as Figure \ref{fig:dp example} shows, and if there exists some utterances not depending on any others, then we assign it to depend on the root. The prediction is divided into two parts. The first one is link prediction: we calculate the existence of relationship between utterances, that is to say, for the $i^{th}$ utterance, we adopt a matrix decomposition by performing SVD over ($F_{i,1}$, $F_{i,2}$, ..., $F_{i,t}$) for significant eigenvector to indicate which utterance it depends on, where $t$ is the max number of utterances in a dialogue. Meanwhile we also use $F_{i,j}$ to predict the kind of the relationship between the $i^{th}$ and $j^{th}$ utterance which is the second part called relationship prediction. We regard these two parts as multi-classification and input the logits into softmax layer and argmax layer to get final answer:
\begin{equation}\label{equation:link}
\begin{split}
L_i&=\textrm{argmax}(\textrm{softmax}(\boldsymbol{W}_{l}[F_{i,1}, F_{i,2}, ..., F_{i,t}])),\\
R_{i,j}&=\textrm{argmax}(\textrm{softmax}(\boldsymbol{W}_{r}F_{i,j})),
\end{split}
\end{equation}
where $L_i$ is the predicted utterance number which the $i^{th}$ depends on, $R_{i,j}$ is the predicted relationship between the $i^{th}$ and $j^{th}$ utterances, $\boldsymbol{W}_{l}$ and $\boldsymbol{W}_{r}$ are the weight matrix.
\begin{figure*}[ht]
\centering\small
    \subfigure[Dialogue example from Ubuntu Chat Corpus]{
    \resizebox{0.9\textwidth}{20mm}{
        \begin{tabular}{|p{0.6cm}p{1.3cm}p{11.5cm}|}
        \hline ~ & \textbf{Speaker} & \textbf{Utterance} \\ 
        $\textrm{U}_0$ & \textit{\textcolor[RGB]{153,50,142}{sipher}} & \textit{\textcolor[RGB]{153,50,142}{bacon5o there 's no `` fixmbr '' with ubuntu .}} \\
        $\textrm{U}_1$ & \textit{\textcolor[RGB]{55,86,34}{Bacon5o}} & \textit{\textcolor[RGB]{55,86,34}{i dont want ubuntu , \textbf{it does n't support my internet} , thus i can not use it}} \\
        $\textrm{U}_2$ & \textit{\textcolor[RGB]{32,119,150}{morfic}} & \textit{\textcolor[RGB]{32,119,150}{my ati has no aiglx support so i ca n't speak for how FILEPATH is}}\\
        $\textrm{U}_3$ & \textit{\textcolor[RGB]{32,119,150}{morfic}} & \textit{\textcolor[RGB]{32,119,150}{your internet is different from mine ? damn bush and his internets !}}\\
        $\textrm{U}_4$ & \textit{\textcolor[RGB]{55,86,34}{Bacon5o}} & \textit{\textcolor[RGB]{55,86,34}{my internet is different why you ask ?}}\\
        $\textrm{U}_5$ & \textit{\textcolor[RGB]{32,119,150}{morfic}} & \textit{\textcolor[RGB]{32,119,150}{your possesive `` my '' on the internet}}\\
        $\textrm{U}_6$ & \textit{\textcolor[RGB]{55,86,34}{Bacon5o}}  & \textit{\textcolor[RGB]{55,86,34}{i use \textbf{a wireless accesspoint} that plugs into my usb}}\\
        \hline
        \end{tabular}}}
    \subfigure[Q\&A example for multi-party dialogue MRC]{
    \resizebox{0.9\textwidth}{10mm}{
        \begin{tabular}{|p{7.2cm}p{7.2cm}|}
        \hline \textbf{Question} & \textbf{Answer} \\ 
        \textit{Why does Bacon5o not want ubuntu ?} & \textit{\textcolor[RGB]{55,86,34}{\textbf{it does n't support my internet}}} \\
        \textit{What does Bacon5o use to plugs into usb ?} & \textit{\textcolor[RGB]{55,86,34}{\textbf{a wireless accesspoint}}} \\
        \textit{What did sipher use ?} & \textit{\textbf{(unanswerable question)}} \\ \hline
        \end{tabular}}}
\caption{\label{fig:mrc example}(a) is an example of dialogue in Molweni. (b) shows questions and corresponding answers based on the dialogue in (a). It is noteworthy that unanswerable questions exist. }
\end{figure*} 
\begin{figure*}[ht]
		\centering
		\includegraphics[width=\textwidth]{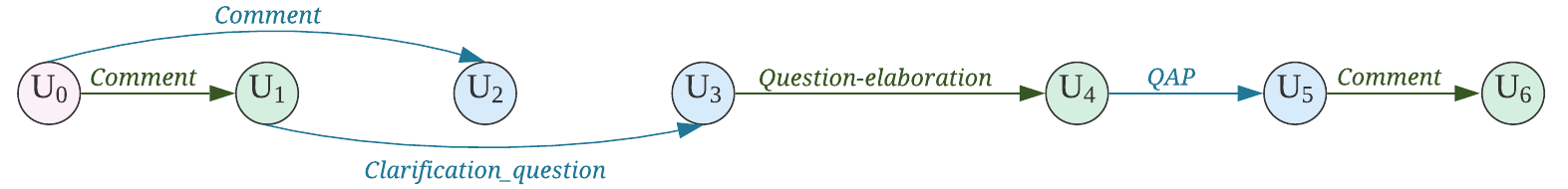}
		\caption{\label{fig:dp example}A dependency tree example for DP task based on the dialogue in Figure \ref{fig:mrc example}. }
\end{figure*}
\subsection{Loss Function}\label{subsec:loss}
Our objective in QA task is to predict the start and end positions for the answers. Assume that there are $K$ tokens in total in the input embedding, then we regard it as multi-classification task with $K$ different labels where one label equals to one position. We firstly use softmax as actived function to normalize the logits, then use cross entropy as loss function to calculate the loss of start and end prediction respectively, and finally average them as total loss of QA task.
\begin{equation}\label{equation:mrc loss}
\begin{split}
\mathbb{L}_m=&-\tfrac{1}{2N}\sum_{n=0}^{N-1}\sum_{k=0}^{K-1}(y_s^{n,k}\log p_s^{n,k}+y_e^{n,k}\log p_e^{n,k}).
\end{split}
\end{equation}
where $\mathbb{L}_m$ is the loss of QA task, $N$ is the batch size, $K$ is the number of labels, $y_s^{n,k}$ equals to one if the answer of the $n^{th}$ sample exactly starts at the $k^{th}$ token or otherwise it equals zero, $p_s^{n,k}$ is the probability of the start position of the $n^{th}$ being predicted to be the $k^{th}$ token and $y_e^{n,k}$ and $p_e^{n,k}$ are similar to $y_s^{n,k}$ and $p_s^{n,k}$ for end position prediction.

For the DP task, the number of relationships is 16 in Molweni as Table \ref{tab:relation types}$\footnote{\label{molweni note}Detailed information can be seen in \newcite{li-etal-2020-molweni}}$ shows, and the max number of utterances in one dialogue is $T$. Then we regard link prediction and relationship prediction as multi-classification with $T+1$ labels and 16 labels respectively, where the additional one label in link prediction is the root. Using cross entropy, the loss function of link prediction $\mathbb{L}_l$ is:
\begin{equation}\label{equation:link loss}
\begin{split}
\mathbb{L}_l=-\tfrac{1}{N}\sum_{n=0}^{N-1}\tfrac{1}{T}(\sum_{t_1=0}^{T-1}\sum_{t_2=0}^{T}y_l^{t_1,t_2}\log p_l^{t_1,t_2}),
\end{split}
\end{equation}
where $y_l^{t_1,t_2}$ equals to one if the $t_1^{th}$ utterance depends on the $t_2^{th}$ one or otherwise it is zero, and $p_l^{t_1,t_2}$ is the probability of the $t_1^{th}$ utterance being predicted to be dependent on the $t_2^{th}$ one. The loss function of relationship prediction $\mathbb{L}_r$ is:
\begin{equation}\label{equation:relationship loss}
\begin{split}
\mathbb{L}_r=-\tfrac{1}{N}\sum_{n=0}^{N-1}\tfrac{1}{T}(\sum_{t=0}^{T-1}\sum_{i=0}^{15}y_r^{t,i}\log p_r^{t,i}),
\end{split}
\end{equation}
where $y_r^{t,i}$ equals to one if the $t^{th}$ utterance depends any other utterance and the relationship is the $i^{th}$ kind or otherwise it is zero, and $p_r^{t,i}$ is the probability of the $t^{th}$ utterance being predicted to be dependent on one utterance and the relationship is the $i^{th}$ one. The loss of DP task $\mathbb{L}_d$ is the sum of $\mathbb{L}_l$ and $\mathbb{L}_r$. Then we add up $\mathbb{L}_d$ with the loss of QA task in Eq.(\ref{equation:mrc loss}) as total loss $\mathbb{L}$ for the joint model.

\section{Experiments}\label{sec:experiments}
\subsection{Molweni Dataset}\label{subsec:dataset}
Molweni dataset \cite{li-etal-2020-molweni} is multi-party dialogue comprehension dataset derived from Ubuntu Chat Corpus \cite{li-etal-2020-molweni}. It has $9,754$ dialogues, $86,042$ utterances and $30,066$ QAPs in total. Among the QAPs, the unanswerable questions account for $14.26\%$. Types of questions are mainly 5W1H which means questions start with \textit{What, Where, When, Who, Why}\textsuperscript{\ref{molweni note}}. For DP task, Molweni has discourse structures for each dialogue and there are $78,245$ discourse relations between utterances in total, among which there are $16$ different kinds as Table \ref{tab:relation types} shows.
\begin{table}[ht]
\renewcommand\arraystretch{1} 
\centering
\small
    \begin{tabular}{p{0.3cm}p{1.5cm}p{0.8cm}<{\raggedleft}|p{0.3cm}p{1.5cm}p{0.8cm}<{\raggedleft}}
    \toprule
    ~ & \textbf{Relation Type} & \textbf{Ratio (\%)} & ~ & \textbf{Relation Type} & \textbf{Ratio (\%)}\\ \midrule
    1 & Comment & $31.7$ &9 &  Explanation & $1.6$ \\
    2 & Clarification question & $24.0$ &10 &  Correction & $1.2$ \\
    3 & QAP & $20.1$ & 11 & Contrast & $1.2$ \\
    4 & Continuation & $6.7$ & 12 & Conditional & $1.0$\\
    5 & Acknowledg- ement & $3.2$ & 13 & Background & $0.4$\\
    6 & Question-elaboration & $3.0$ & 14 & Narration & $0.3$\\
    7 & Result & $2.6$ & 15 & Alternation & $0.2$ \\
    8 & Elaboration & $2.2$ & 16 & Parallel & $0.2$ \\
    \bottomrule
    \end{tabular}
\caption{\label{tab:relation types}The kinds of discourse relations}
\end{table}

Molweni uses both manual check and programmatic check to guarantee its reliability. The Fleiss kappa is $0.91$ for link annotation and $0.56$ for link\&relation annotation which indicates that Molweni has high reliability and consistency.

\subsection{Metrics}\label{subsec:metrics}
Following \newcite{li-etal-2020-molweni}, we use F1 score and exact match (EM) as metrics in QA task. For DP task, we use micro F1 score to judge the link prediction and relationship prediction respectively. For relationship prediction, only when the link and relationship are both correct, it will be counted as positive.
\subsection{Detailed Settings}\label{subsec:detailed settings}
We use three different settings of BERT \cite{devlin-etal-2019-bert} as the PrLM: BERT-base-uncased ($\textrm{BERT}_{base}$), BERT-large-uncased ($\textrm{BERT}_{large}$) and BERT-large-uncased-whole-word-masking ($\textrm{BERT}_{wwm}$).
The hidden size of each model is $768$, $1024$, and $1024$ respectively. The max sequence length is 512 in tokens, and the max utterance number per dialogue is 14 according to \newcite{li-etal-2020-molweni}. Based on the results on the dev set, we set the learning rate to $5e^{-5}$ for $\textrm{BERT}_{base}$, $3e^{-5}$ for $\textrm{BERT}_{large}$, $3e^{-5}$ for $\textrm{BERT}_{wwm}$, and set the dropout rate of DP task to 0.4 for $\textrm{BERT}_{base}$, 0.4 for $\textrm{BERT}_{large}$, 0.1 for $\textrm{BERT}_{wwm}$. 

In the fine-tuning stage, we train all the models for 2 epochs. We try three different values 0.5, 1, and 2 as the ratio of $\mathbb{L}_m$ to $\mathbb{L}_d$. We finally set the ratio to 1 which gets the best result.
\subsection{Results}\label{subsec:results}
The results of our experiments together with public results and human performance are in Table \ref{tab:results}. We see that compared to QA-only model, the results of QA in multi-tasking model make a progress, and this may also apply to the results of DP task. It shows that our joint model indeed leads to a mutual promotion. Furthermore, we compare our results with the benchmark of \newcite{li-etal-2020-molweni} in Table \ref{tab:results}, showing that our model achieves new state-of-the-art in both QA and DP task.
\begin{table*}[ht]
\renewcommand\arraystretch{1} 
    \centering
    \small
    \begin{tabular}{l|lcccc}
    \toprule
         \multicolumn{2}{l}{\multirow{2}*{Method}} & \multicolumn{2}{c}{QA} & \multicolumn{2}{c}{DP} \\ \cline{3-6}
    \specialrule{0em}{1pt}{2pt}
         \multicolumn{2}{c}{~} & F1(\%) & EM(\%) & Link(\%) & Relationship(\%) \\ \midrule
         \multicolumn{2}{l}{Human performance} & 80.2 & 64.3 & - & - \\ 
         \multicolumn{2}{l}{Deep sequential(\newcite{li-etal-2020-molweni})} & - & - & 78.1 & 54.8 \\ \hline
         \multirow{4}*{$\textrm{BERT}_{base}$} & \newcite{li-etal-2020-molweni} & 58.0 & 45.3 & - & -  \\
         ~ & QA-only & 59.2 & 46.2 & - & - \\
         ~ & DP-only & - & - & 73.9 & 56.1 \\
         ~ & Multi-task & 61.3 (+2.1) & 47.1 (+0.9) & 75.9 (+2.0) & 56.2 (+0.1) \\ \hline
         \multirow{4}*{$\textrm{BERT}_{large}$} & \newcite{li-etal-2020-molweni} & 65.5 & 51.8 & - & -\\
         ~ & QA-only & 64.0 & 49.6 & - & - \\
         ~ & DP-only & - & - & 81.0 & 61.5 \\
         ~ & Multi-task & 64.9 (+0.9) & 50.6 (+1.0) & 82.1 (+1.1) & 62.0 (+0.5) \\ \hline
         \multirow{4}*{$\textrm{BERT}_{wwm}$} & \newcite{li-etal-2020-molweni} & 67.7 & 54.7 & - & - \\
         ~ & QA-only & 67.5 & 53.8  & - & - \\
         ~ & DP-only & - & - & 86.6 & 64.9 \\
         ~ & Multi-task & \textbf{68.4} (+0.9) & \textbf{54.9} (+1.1) & \textbf{88.1} (+1.5) & \textbf{66.9} (+2.0) \\
        \bottomrule
    \end{tabular}
    \caption{Results on Molweni dataset. Results except ours are from \newcite{li-etal-2020-molweni}.}
    \label{tab:results}
\end{table*}

Besides, by analysing the performances of our joint model under different parameters, we discover that the results of the two tasks are closely linked to each other. For example, when the DP task in our model is overfitting or even not convergent at all, the performance of QA task will also decrease to a certain extent which verifies the close correlations between QA and DP.

Additionally, compared to the time cost per iteration of single task model, the joint model does not take extra time. For DP task that shares the dataset and text features with QA task, it only needs an additional fully connected layer and a softmax layer as an actived function whose time cost is negligible. We combine the loss of DP and QA together to feed back to the model, so during the phases of feature extraction and back propagation, there will not be any extra cost.

\section{Analysis}\label{subsec:analysis}
\subsection{DP Improvement Analysis}\label{subsubsec:dp analysis}
Compared to single DP model, multi-tasking model can better parse the discourse structure. The possible reason is that QA task pays attention to extracting answer spans which requires the capacity to obtain salient information from utterances. Thus this relieves the problem of long distance dependency. This capacity also helps the DP task to resist the noise of long texts, and may have a positive impact on parsing nonadjacent utterance relationship.

To verify our speculation, we further extract and analyze the predictions of nonadjacent utterance relationship which is a relatively difficult part in DP task. We calculate the F1 scores of these predictions on both multi-tasking and single DP models on $\textrm{BERT}_{wwm}$. For link prediction, the F1 score of multi-tasking model is $56.2\%$, which is $1.4\%$ higher than that of single DP task. For relationship prediction, the F1 score of multi-tasking model is $37.3\%$, which outperforms single DP task by $3.0\%$. We see that there are noticeable increases in both link and relationship predictions, which proves that with the help of QA task, DP task can better resist the noise of complex texts and predict nonadjacent utterance relationship more precisely.

\subsection{QA Improvement Analysis}\label{subsec:QA improvement analysis}
\begin{figure}[ht]
		\centering
		\includegraphics[width=0.45\textwidth]{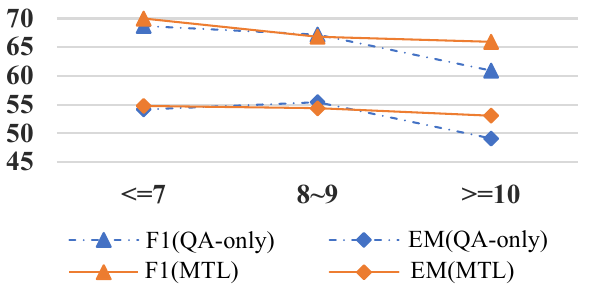}
		\caption{\label{fig:dialogue length}The results of dialogues with different numbers of utterances (on $\textrm{BERT}_{wwm}$).}
\end{figure}

We divide the test set into three parts based on dialogue length: dialogues with less than or equal to 7 utterances (account for 40\%), dialogues with 8 or 9 utterances (account for 31\%) and dialogues with more than or equal to 10 utterances (account for 29\%). We evaluate the QA-only model and MTL model respectively on these three subsets to further explore the impact of DP task on QA task. The results are shown in Figure \ref{fig:dialogue length}. It shows that the QA-only performance on long dialogues is obviously worse than short ones. The reason could be the QA-only model can only obtain limited context information. When the distance between utterances is far, it can no longer pay enough attention to the relationship of these utterances which might actually be tightly interconnected.

It can be observed from Figure \ref{fig:dialogue length} that though the performances of MTL and QA-only on short dialogues have little difference, the MTL model can distinctly better handle longer dialogues. The results of MTL on long dialogues drop little compared to short dialogues, showing that MTL might benefit from the DP task which pays equal attention to related utterances even though they are far apart.

\subsection{Case Analysis}\label{subsubsec:case analysis} 
To further explore the effect of discourse structures on multi-party dialogue MRC, we compare all the QAPs predicted by multi-tasking model and single QA model respectively (on $\textrm{BERT}_{wwm}$). We intentionally fetch the answerable questions which are answered correctly on joint model while wrongly on single QA model. There are 99 such QAPs in the test set. Through artificial judging, we find 58 in 99 QAPs 
which confirms the help of DP to QA. For example, there is the following dialogue:
\\  \\
\vspace{-1cm}
\begin{itemize}[leftmargin=0.3cm]
\item[]
	 \textit{Suikwan:}  \textit{``do you know where i can get the linux drivers ?''} \\
	 \textit{arkady:}  \textit{``apparently that is `` old and unsupported '' by d-link , and they do n't have linux drivers''} \\
	 \textit{arkady:}  \textit{``you can use ndiswrapper to wrap the windows drivers , then''}\\
\end{itemize}
\vspace{-0.5cm}

For the question \textit{Where to get the linux drivers}, the joint model answer is 
\textit{use ndiswrapper to wrap the windows drivers} which is exactly the same with gold answer while the answer of single QA model is \textit{by d-link}. Owing to the discourse information, joint model puts more emphasis on the third turn because it captures the \textit{QAP} relationship between the first and third utterances. By contrast, the QA-only task pays attention to traditional context, so it naturally extracts the answer from the adjacent utterance. These 58 in 99 cases are strong evidence for the importance of discourse parsing in multi-party dialogue MRC.

\begin{figure}[ht]
		\centering
		\includegraphics[width=0.45\textwidth]{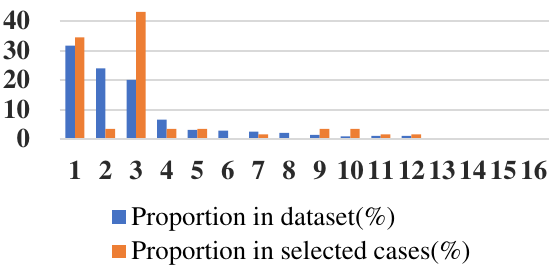}
		\caption{\label{fig:relationships proportion}The proportion of main discourse relationships in Molweni dataset and the cases we choose. The relationship types name correspond to the types in Table \ref{tab:relation types}.}
\end{figure}
To explore the detailed effects of different relationships, we calculated the proportion of each relationship in the 58 cases we choose. Figure \ref{fig:relationships proportion} shows the result. We see that \textit{QAP} accounts for a large proportion and makes a significant contribution to QA task. By contrast, \textit{Clarification question} is not so important for QA.
This inspires us that annotating the main contributive relationships like \textit{QAP} precisely is very helpful to multi-party dialogue comprehension.

\subsection{Error Analysis}\label{subsubsec:error analysis}
\begin{figure}[ht]
		\centering
		\includegraphics[width=0.45\textwidth]{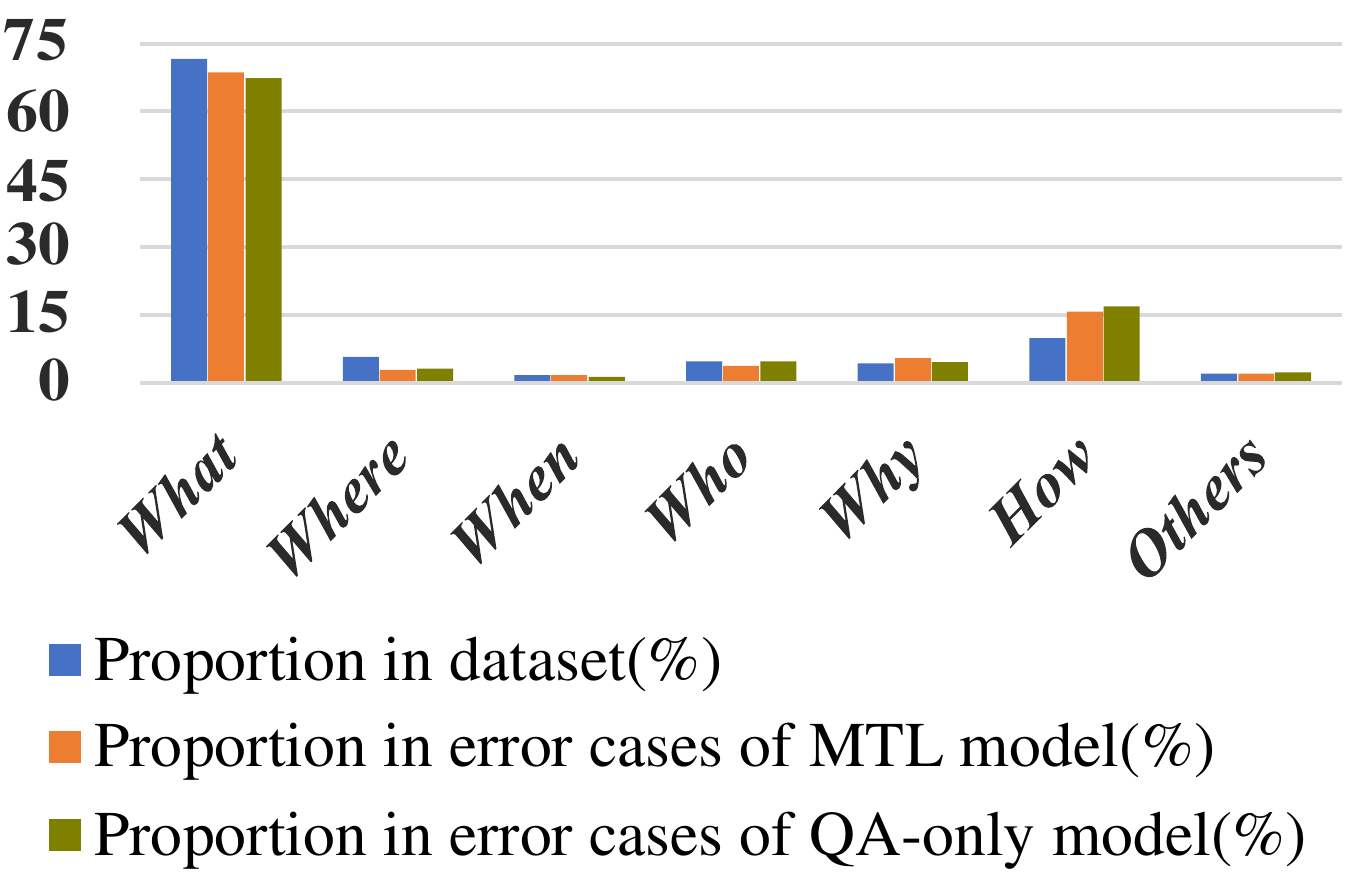}
		\caption{\label{fig:question proportion}The proportion of question types in Molweni dataset, in the error cases of multi-tasking model and single QA model.}
\end{figure}
In order to explore the potential improvement room, we statistically analyze the error cases of both single QA model and multi-tasking model. As shown in Figure \ref{fig:question proportion}, we calculate the proportion of each kind of questions in the error cases of these two models. Questions start with \textit{what} account for the majority which is not surprising because most of the Molweni dataset is \textit{what}-leading questions. It is worth noting that multi-tasking can better answer \textit{who}-leading questions. The possible reason is that \textit{who}-leading questions like \textit{Who answered BrandonBolton ?} focuses on the relationship between speakers which is exactly what the discourse structures are for.

It is also distinct in Figure \ref{fig:question proportion} that \textit{how}-leading questions are challenging for both single QA and multi-tasking model. We attribute this difficulty to the too flexible and too diverse for the usage of \textit{how}-leading questions. Compared to \textit{how}, questions start with other adverbs such as \textit{where}, \textit{when} and other interrogative pronouns are more concrete and easier. This inspires us that syntactic analysis may has an impact on \textit{how}-leading questions which worth a try.

\section{Conclusion}\label{sec:conclusion}
In this paper, we are motivated to investigate the correlationship between QA and DP tasks. To this end, we propose the first multi-task model for jointly performing QA and DP on one multi-party dialogue MRC to blend the discourse structures with answer extraction. Results indicate that our joint model indeed improves the performance of both QA and DP tasks, which proves that there exists a strong and positive correlationship between these two tasks. A series of analyses are conducted to explore the contributing factors. For cases that the dialogue datasets might not have the corresponding discourse annotations, it is possible to apply off-the-shelf dialogue discourse parsing tools to obtain the discourse relationships  \cite{ouyang2020dialogue}, which is left for future work. In addition, it would be interesting to investigate graph networks to model complex QA based on discourse structures and improve the reasoning ability of dialogue systems.

\bibliographystyle{acl}

\newpage
\begin{thebibliography}{}

\bibitem[\protect\citename{Afantenos \bgroup et al.\egroup
  }2015]{afantenos2015discourse}
Stergos Afantenos, Eric Kow, Nicholas Asher, and J{\'e}r{\'e}my Perret.
\newblock 2015.
\newblock Discourse parsing for multi-party chat dialogues.
\newblock In {\em Proceedings of the 2015 Conference on Empirical Methods in
  Natural Language Processing}, pages 928--937, Lisbon, Portugal. Association
  for Computational Linguistics.

\bibitem[\protect\citename{Allen \bgroup et al.\egroup }1994]{allen1994trains}
James~F Allen, Lenhart~K Schubert, George~M Ferguson, Peter~A Heeman, and
  Chung~Hee Hwang.
\newblock 1994.
\newblock The trains project: A case study in building a conversational
  planning agent.
\newblock Technical report, ROCHESTER UNIV NY DEPT OF COMPUTER SCIENCE.

\bibitem[\protect\citename{Asher \bgroup et al.\egroup
  }2016]{asher2016discourse}
Nicholas Asher, Julie Hunter, Mathieu Morey, Benamara Farah, and Stergos
  Afantenos.
\newblock 2016.
\newblock Discourse structure and dialogue acts in multiparty dialogue: the
  {STAC} corpus.
\newblock In {\em Proceedings of the Tenth International Conference on Language
  Resources and Evaluation ({LREC}'16)}, pages 2721--2727, Portoro{\v{z}},
  Slovenia. European Language Resources Association (ELRA).

\bibitem[\protect\citename{Bai and Zhao}2018]{bai2018deep}
Hongxiao Bai and Hai Zhao.
\newblock 2018.
\newblock Deep enhanced representation for implicit discourse relation
  recognition.
\newblock In {\em Proceedings of the 27th International Conference on
  Computational Linguistics}, pages 571--583, Santa Fe, New Mexico, USA.
  Association for Computational Linguistics.

\bibitem[\protect\citename{Braud \bgroup et al.\egroup }2017]{braud2017cross}
Chlo{\'e} Braud, Maximin Coavoux, and Anders S{\o}gaard.
\newblock 2017.
\newblock Cross-lingual {RST} discourse parsing.
\newblock In {\em Proceedings of the 15th Conference of the {E}uropean Chapter
  of the Association for Computational Linguistics: Volume 1, Long Papers},
  pages 292--304, Valencia, Spain. Association for Computational Linguistics.

\bibitem[\protect\citename{Cai and Zhao}2017]{cai2017pair}
Deng Cai and Hai Zhao.
\newblock 2017.
\newblock Pair-aware neural sentence modeling for implicit discourse relation
  classification.
\newblock In {\em International Conference on Industrial, Engineering and Other
  Applications of Applied Intelligent Systems}, pages 458--466. Springer.

\bibitem[\protect\citename{Cambria \bgroup et al.\egroup }2013]{cambria2013new}
Erik Cambria, Bj{\"o}rn Schuller, Yunqing Xia, and Catherine Havasi.
\newblock 2013.
\newblock New avenues in opinion mining and sentiment analysis.
\newblock {\em IEEE Intelligent systems}, 28(2):15--21.

\bibitem[\protect\citename{Chai and Jin}2004]{chai2004discourse}
Joyce~Y. Chai and Rong Jin.
\newblock 2004.
\newblock Discourse structure for context question answering.
\newblock In {\em Proceedings of the Workshop on Pragmatics of Question
  Answering at {HLT}-{NAACL} 2004}, pages 23--30, Boston, Massachusetts, USA.
  Association for Computational Linguistics.

\bibitem[\protect\citename{Choi \bgroup et al.\egroup }2018]{choi2018quac}
Eunsol Choi, He~He, Mohit Iyyer, Mark Yatskar, Wen-tau Yih, Yejin Choi, Percy
  Liang, and Luke Zettlemoyer.
\newblock 2018.
\newblock {Q}u{AC}: Question answering in context.
\newblock In {\em Proceedings of the 2018 Conference on Empirical Methods in
  Natural Language Processing}, pages 2174--2184, Brussels, Belgium.
  Association for Computational Linguistics.

\bibitem[\protect\citename{Clark \bgroup et al.\egroup
  }2020]{Clark2020ELECTRA:}
Kevin Clark, Minh{-}Thang Luong, Quoc~V. Le, and Christopher~D. Manning.
\newblock 2020.
\newblock {ELECTRA:} pre-training text encoders as discriminators rather than
  generators.
\newblock In {\em 8th International Conference on Learning Representations,
  {ICLR} 2020, Addis Ababa, Ethiopia, April 26-30, 2020}. OpenReview.net.

\bibitem[\protect\citename{Devlin \bgroup et al.\egroup
  }2019]{devlin-etal-2019-bert}
Jacob Devlin, Ming-Wei Chang, Kenton Lee, and Kristina Toutanova.
\newblock 2019.
\newblock {BERT}: Pre-training of deep bidirectional transformers for language
  understanding.
\newblock In {\em Proceedings of the 2019 Conference of the North {A}merican
  Chapter of the Association for Computational Linguistics: Human Language
  Technologies, Volume 1 (Long and Short Papers)}, pages 4171--4186,
  Minneapolis, Minnesota. Association for Computational Linguistics.

\bibitem[\protect\citename{Gao \bgroup et al.\egroup
  }2020]{gao-etal-2020-discern}
Yifan Gao, Chien-Sheng Wu, Jingjing Li, Shafiq Joty, Steven~C.H. Hoi, Caiming
  Xiong, Irwin King, and Michael Lyu.
\newblock 2020.
\newblock Discern: Discourse-aware entailment reasoning network for
  conversational machine reading.
\newblock In {\em Proceedings of the 2020 Conference on Empirical Methods in
  Natural Language Processing (EMNLP)}, pages 2439--2449, Online. Association
  for Computational Linguistics.

\bibitem[\protect\citename{Hermann \bgroup et al.\egroup
  }2015]{hermann2015teaching}
Karl~Moritz Hermann, Tom{\'{a}}s Kocisk{\'{y}}, Edward Grefenstette, Lasse
  Espeholt, Will Kay, Mustafa Suleyman, and Phil Blunsom.
\newblock 2015.
\newblock Teaching machines to read and comprehend.
\newblock In Corinna Cortes, Neil~D. Lawrence, Daniel~D. Lee, Masashi Sugiyama,
  and Roman Garnett, editors, {\em Advances in Neural Information Processing
  Systems 28: Annual Conference on Neural Information Processing Systems 2015,
  December 7-12, 2015, Montreal, Quebec, Canada}, pages 1693--1701.

\bibitem[\protect\citename{Hill \bgroup et al.\egroup
  }2016]{hill2015goldilocks}
Felix Hill, Antoine Bordes, Sumit Chopra, and Jason Weston.
\newblock 2016.
\newblock The goldilocks principle: Reading children's books with explicit
  memory representations.
\newblock In Yoshua Bengio and Yann LeCun, editors, {\em 4th International
  Conference on Learning Representations, {ICLR} 2016, San Juan, Puerto Rico,
  May 2-4, 2016, Conference Track Proceedings}.

\bibitem[\protect\citename{Jia \bgroup et al.\egroup }2020]{jia2020multi}
Qi~Jia, Yizhu Liu, Siyu Ren, Kenny Zhu, and Haifeng Tang.
\newblock 2020.
\newblock Multi-turn response selection using dialogue dependency relations.
\newblock In {\em Proceedings of the 2020 Conference on Empirical Methods in
  Natural Language Processing (EMNLP)}, pages 1911--1920, Online. Association
  for Computational Linguistics.

\bibitem[\protect\citename{Joty \bgroup et al.\egroup }2015]{joty2015codra}
Shafiq Joty, Giuseppe Carenini, and Raymond~T. Ng.
\newblock 2015.
\newblock {CODRA}: A novel discriminative framework for rhetorical analysis.
\newblock {\em Computational Linguistics}, 41(3):385--435.

\bibitem[\protect\citename{Lai \bgroup et al.\egroup }2017]{lai2017race}
Guokun Lai, Qizhe Xie, Hanxiao Liu, Yiming Yang, and Eduard Hovy.
\newblock 2017.
\newblock {RACE}: Large-scale {R}e{A}ding comprehension dataset from
  examinations.
\newblock In {\em Proceedings of the 2017 Conference on Empirical Methods in
  Natural Language Processing}, pages 785--794, Copenhagen, Denmark.
  Association for Computational Linguistics.

\bibitem[\protect\citename{Li \bgroup et al.\egroup }2016]{li2016discourse}
Qi~Li, Tianshi Li, and Baobao Chang.
\newblock 2016.
\newblock Discourse parsing with attention-based hierarchical neural networks.
\newblock In {\em Proceedings of the 2016 Conference on Empirical Methods in
  Natural Language Processing}, pages 362--371, Austin, Texas. Association for
  Computational Linguistics.

\bibitem[\protect\citename{Li \bgroup et al.\egroup
  }2020]{li-etal-2020-molweni}
Jiaqi Li, Ming Liu, Min-Yen Kan, Zihao Zheng, Zekun Wang, Wenqiang Lei, Ting
  Liu, and Bing Qin.
\newblock 2020.
\newblock Molweni: A challenge multiparty dialogues-based machine reading
  comprehension dataset with discourse structure.
\newblock In {\em Proceedings of the 28th International Conference on
  Computational Linguistics}, pages 2642--2652, Barcelona, Spain (Online).
  International Committee on Computational Linguistics.

\bibitem[\protect\citename{Li \bgroup et al.\egroup
  }2021]{DBLP:journals/corr/abs-2104-12377}
Jiaqi Li, Ming Liu, Zihao Zheng, Heng Zhang, Bing Qin, Min{-}Yen Kan, and Ting
  Liu.
\newblock 2021.
\newblock Dadgraph: {A} discourse-aware dialogue graph neural network for
  multiparty dialogue machine reading comprehension.
\newblock {\em CoRR}, abs/2104.12377.

\bibitem[\protect\citename{Liu and Lapata}2017]{liu2017learning}
Yang Liu and Mirella Lapata.
\newblock 2017.
\newblock Learning contextually informed representations for linear-time
  discourse parsing.
\newblock In {\em Proceedings of the 2017 Conference on Empirical Methods in
  Natural Language Processing}, pages 1289--1298, Copenhagen, Denmark.
  Association for Computational Linguistics.

\bibitem[\protect\citename{Mann and Thompson}1988]{mann1988rhetorical}
William~C Mann and Sandra~A Thompson.
\newblock 1988.
\newblock Rhetorical structure theory: Toward a functional theory of text
  organization.
\newblock {\em Text}, 8(3):243--281.

\bibitem[\protect\citename{Mihaylov and Frank}2019]{mihaylov2019discourse}
Todor Mihaylov and Anette Frank.
\newblock 2019.
\newblock Discourse-aware semantic self-attention for narrative reading
  comprehension.
\newblock In {\em Proceedings of the 2019 Conference on Empirical Methods in
  Natural Language Processing and the 9th International Joint Conference on
  Natural Language Processing (EMNLP-IJCNLP)}, pages 2541--2552, Hong Kong,
  China. Association for Computational Linguistics.

\bibitem[\protect\citename{Miltsakaki \bgroup et al.\egroup
  }2004]{prasad2008penn}
Eleni Miltsakaki, Rashmi Prasad, Aravind Joshi, and Bonnie Webber.
\newblock 2004.
\newblock The {P}enn {D}iscourse {T}reebank.
\newblock In {\em Proceedings of the Fourth International Conference on
  Language Resources and Evaluation ({LREC}{'}04)}, Lisbon, Portugal. European
  Language Resources Association (ELRA).

\bibitem[\protect\citename{Nejat \bgroup et al.\egroup
  }2017]{nejat2017exploring}
Bita Nejat, Giuseppe Carenini, and Raymond Ng.
\newblock 2017.
\newblock Exploring joint neural model for sentence level discourse parsing and
  sentiment analysis.
\newblock In {\em Proceedings of the 18th Annual {SIG}dial Meeting on Discourse
  and Dialogue}, pages 289--298, Saarbr{\"u}cken, Germany. Association for
  Computational Linguistics.

\bibitem[\protect\citename{Ouyang \bgroup et al.\egroup
  }2021]{ouyang2020dialogue}
Siru Ouyang, Zhuosheng Zhang, and Hai Zhao.
\newblock 2021.
\newblock Dialogue graph modeling for conversational machine reading.
\newblock In {\em Findings of the Association for Computational Linguistics:
  ACL-IJCNLP 2021}, pages 3158--3169, Online. Association for Computational
  Linguistics.

\bibitem[\protect\citename{Perez and Liu}2017]{perez2017dialog}
Julien Perez and Fei Liu.
\newblock 2017.
\newblock Dialog state tracking, a machine reading approach using memory
  network.
\newblock In {\em Proceedings of the 15th Conference of the {E}uropean Chapter
  of the Association for Computational Linguistics: Volume 1, Long Papers},
  pages 305--314, Valencia, Spain. Association for Computational Linguistics.

\bibitem[\protect\citename{Qin \bgroup et al.\egroup }2017]{qin2017adversarial}
Lianhui Qin, Zhisong Zhang, Hai Zhao, Zhiting Hu, and Eric Xing.
\newblock 2017.
\newblock Adversarial connective-exploiting networks for implicit discourse
  relation classification.
\newblock In {\em Proceedings of the 55th Annual Meeting of the Association for
  Computational Linguistics (Volume 1: Long Papers)}, pages 1006--1017,
  Vancouver, Canada. Association for Computational Linguistics.

\bibitem[\protect\citename{Rajpurkar \bgroup et al.\egroup
  }2016]{rajpurkar2016squad}
Pranav Rajpurkar, Jian Zhang, Konstantin Lopyrev, and Percy Liang.
\newblock 2016.
\newblock {SQ}u{AD}: 100,000+ questions for machine comprehension of text.
\newblock In {\em Proceedings of the 2016 Conference on Empirical Methods in
  Natural Language Processing}, pages 2383--2392, Austin, Texas. Association
  for Computational Linguistics.

\bibitem[\protect\citename{Rajpurkar \bgroup et al.\egroup
  }2018]{Rajpurkar2018Know}
Pranav Rajpurkar, Robin Jia, and Percy Liang.
\newblock 2018.
\newblock Know what you don{'}t know: Unanswerable questions for {SQ}u{AD}.
\newblock In {\em Proceedings of the 56th Annual Meeting of the Association for
  Computational Linguistics (Volume 2: Short Papers)}, pages 784--789,
  Melbourne, Australia. Association for Computational Linguistics.

\bibitem[\protect\citename{Reddy \bgroup et al.\egroup }2019]{reddy2019coqa}
Siva Reddy, Danqi Chen, and Christopher~D. Manning.
\newblock 2019.
\newblock {C}o{QA}: A conversational question answering challenge.
\newblock {\em Transactions of the Association for Computational Linguistics},
  7:249--266.

\bibitem[\protect\citename{Seo \bgroup et al.\egroup
  }2017]{seo2018bidirectional}
Min~Joon Seo, Aniruddha Kembhavi, Ali Farhadi, and Hannaneh Hajishirzi.
\newblock 2017.
\newblock Bidirectional attention flow for machine comprehension.
\newblock In {\em 5th International Conference on Learning Representations,
  {ICLR} 2017, Toulon, France, April 24-26, 2017, Conference Track
  Proceedings}. OpenReview.net.

\bibitem[\protect\citename{Shi and Huang}2019]{shi2019deep}
Zhouxing Shi and Minlie Huang.
\newblock 2019.
\newblock A deep sequential model for discourse parsing on multi-party
  dialogues.
\newblock In {\em The Thirty-Third {AAAI} Conference on Artificial
  Intelligence, {AAAI} 2019, The Thirty-First Innovative Applications of
  Artificial Intelligence Conference, {IAAI} 2019, The Ninth {AAAI} Symposium
  on Educational Advances in Artificial Intelligence, {EAAI} 2019, Honolulu,
  Hawaii, USA, January 27 - February 1, 2019}, pages 7007--7014. {AAAI} Press.

\bibitem[\protect\citename{Sun \bgroup et al.\egroup }2019]{sun2019dream}
Kai Sun, Dian Yu, Jianshu Chen, Dong Yu, Yejin Choi, and Claire Cardie.
\newblock 2019.
\newblock {DREAM}: A challenge data set and models for dialogue-based reading
  comprehension.
\newblock {\em Transactions of the Association for Computational Linguistics},
  7:217--231.

\bibitem[\protect\citename{Takanobu \bgroup et al.\egroup
  }2018]{takanobu2018weakly}
Ryuichi Takanobu, Minlie Huang, Zhongzhou Zhao, Feng{-}Lin Li, Haiqing Chen,
  Xiaoyan Zhu, and Liqiang Nie.
\newblock 2018.
\newblock A weakly supervised method for topic segmentation and labeling in
  goal-oriented dialogues via reinforcement learning.
\newblock In J{\'{e}}r{\^{o}}me Lang, editor, {\em Proceedings of the
  Twenty-Seventh International Joint Conference on Artificial Intelligence,
  {IJCAI} 2018, July 13-19, 2018, Stockholm, Sweden}, pages 4403--4410.
  ijcai.org.

\bibitem[\protect\citename{Verberne \bgroup et al.\egroup
  }2007]{verberne2007evaluating}
Suzan Verberne, Lou Boves, Nelleke Oostdijk, and Peter-Arno Coppen.
\newblock 2007.
\newblock Evaluating discourse-based answer extraction for why-question
  answering.
\newblock In {\em Proceedings of the 30th annual international ACM SIGIR
  conference on Research and development in information retrieval}, pages
  735--736.

\bibitem[\protect\citename{Wang \bgroup et al.\egroup }2017]{wang2017two}
Yizhong Wang, Sujian Li, and Houfeng Wang.
\newblock 2017.
\newblock A two-stage parsing method for text-level discourse analysis.
\newblock In {\em Proceedings of the 55th Annual Meeting of the Association for
  Computational Linguistics (Volume 2: Short Papers)}, pages 184--188,
  Vancouver, Canada. Association for Computational Linguistics.

\bibitem[\protect\citename{Yang and Li}2018]{yang2018scidtb}
An~Yang and Sujian Li.
\newblock 2018.
\newblock {S}ci{DTB}: Discourse dependency {T}ree{B}ank for scientific
  abstracts.
\newblock In {\em Proceedings of the 56th Annual Meeting of the Association for
  Computational Linguistics (Volume 2: Short Papers)}, pages 444--449,
  Melbourne, Australia. Association for Computational Linguistics.

\bibitem[\protect\citename{Yu \bgroup et al.\egroup }2018a]{wei2018fast}
Adams~Wei Yu, David Dohan, Minh{-}Thang Luong, Rui Zhao, Kai Chen, Mohammad
  Norouzi, and Quoc~V. Le.
\newblock 2018a.
\newblock Qanet: Combining local convolution with global self-attention for
  reading comprehension.
\newblock In {\em 6th International Conference on Learning Representations,
  {ICLR} 2018, Vancouver, BC, Canada, April 30 - May 3, 2018, Conference Track
  Proceedings}. OpenReview.net.

\bibitem[\protect\citename{Yu \bgroup et al.\egroup }2018b]{yu2018transition}
Nan Yu, Meishan Zhang, and Guohong Fu.
\newblock 2018b.
\newblock Transition-based neural {RST} parsing with implicit syntax features.
\newblock In {\em Proceedings of the 27th International Conference on
  Computational Linguistics}, pages 559--570, Santa Fe, New Mexico, USA.
  Association for Computational Linguistics.

\bibitem[\protect\citename{Zhang \bgroup et al.\egroup
  }2020a]{zhang2020retrospective}
Zhuosheng Zhang, Junjie Yang, and Hai Zhao.
\newblock 2020a.
\newblock Retrospective reader for machine reading comprehension.

\bibitem[\protect\citename{Zhang \bgroup et al.\egroup
  }2020b]{zhang2020machine}
Zhuosheng Zhang, Hai Zhao, and Rui Wang.
\newblock 2020b.
\newblock Machine reading comprehension: The role of contextualized language
  models and beyond.
\newblock {\em ArXiv preprint}, abs/2005.06249.

\end{thebibliography}

\begin{thebibliography}{}

\bibitem[\protect\citename{Aho and Ullman}1972]{Aho:72}
Alfred~V. Aho and Jeffrey~D. Ullman.
\newblock 1972.
\newblock {\em The Theory of Parsing, Translation and Compiling}, volume~1.
\newblock Prentice-{Hall}, Englewood Cliffs, NJ.

\bibitem[\protect\citename{{American Psychological Association}}1983]{APA:83}
{American Psychological Association}.
\newblock 1983.
\newblock {\em Publications Manual}.
\newblock American Psychological Association, Washington, DC.

\bibitem[\protect\citename{{Association for Computing Machinery}}1983]{ACM:83}
{Association for Computing Machinery}.
\newblock 1983.
\newblock {\em Computing Reviews}, 24(11):503--512.

\bibitem[\protect\citename{Chandra \bgroup et al.\egroup }1981]{Chandra:81}
Ashok~K. Chandra, Dexter~C. Kozen, and Larry~J. Stockmeyer.
\newblock 1981.
\newblock Alternation.
\newblock {\em Journal of the Association for Computing Machinery},
  28(1):114--133.

\bibitem[\protect\citename{Gusfield}1997]{Gusfield:97}
Dan Gusfield.
\newblock 1997.
\newblock {\em Algorithms on Strings, Trees and Sequences}.
\newblock Cambridge University Press, Cambridge, UK.

\end{thebibliography}

\end{document}